\pdfoutput=1

\documentclass[11pt]{article}

\PassOptionsToPackage{table}{xcolor}
\usepackage{acl}

\usepackage{times}
\usepackage{latexsym}

\usepackage[T1]{fontenc}

\usepackage[utf8]{inputenc}

\usepackage{microtype}

\usepackage{inconsolata}

\usepackage{graphicx}

\usepackage{url}
\usepackage{fontawesome5}

\usepackage{graphicx}
\usepackage{rotating}
\usepackage{multirow}
\usepackage{booktabs}
\usepackage{adjustbox}
\usepackage{amssymb}
\usepackage{makecell}
\usepackage{amsmath}
\newcommand{\cmark}{\checkmark}

\newcommand{\thinmidrule}{%
  \noalign{\global\setlength{\belowrulesep}{0pt}}%
  \midrule
  \noalign{\global\setlength{\belowrulesep}{0.65ex}}
}

\newcommand*{\twoelementtable}[3][l]%
{%  
    \begin{tabular}[t]{@{}#1@{}}%
        #2\tabularnewline
        #3%
    \end{tabular}%
}
\usepackage{xparse}  
\NewDocumentCommand{\rot}{O{45} O{1.5em} m}{%
  \makebox[#2][l]{\rotatebox{#1}{\parbox{1.5cm}{\centering #3}}}%
}

\title{PIIvot: A Lightweight NLP Anonymization Framework for Question-Anchored Tutoring Dialogues}

\author{Matthew Zent \quad Digory Smith \quad Simon Woodhead \\
  Eedi \\
  \texttt{matthew.zent@eedi.co.uk}}

\begin{document}
\maketitle
\begin{abstract}
Personally identifiable information (PII) anonymization is a high-stakes task that poses a barrier to many open-science data sharing initiatives.
While PII identification has made large strides in recent years, in practice, error thresholds and the recall/precision trade-off still limit the uptake of these anonymization pipelines.
We present PIIvot, a lighter-weight framework for PII anonymization that leverages knowledge of the data context to simplify the PII detection problem.
To demonstrate its effectiveness, we also contribute QATD\textsubscript{2k}, the largest open-source real-world tutoring dataset of its kind, to support the demand for quality educational dialogue data.
\end{abstract}

\noindent
\faGithub\ \url{https://github.com/Eedi/PIIvot} \\
\url{https://huggingface.co/datasets/Eedi/Question-Anchored-Tutoring-Dialogues-2k}

\section{Introduction and Related Work}
\label{sec:intro}

Over the past 10 years, we've seen widespread adoption and growth of education technology inside and outside the classroom~\citep{escueta_education_2017, manal_edutech_nodate, manna_edutech_2022}.
Understanding and improving affective learning strategies continues to be one of computing's primary contributions to education research ~\citep{mandalapu_understanding_2019}.
Among these advancements, high-dosage online tutoring has emerged as a particularly effective intervention to enhance student learning outcomes~\citep{carlana2024apart, gortazar_online_2024}, but faces barriers to equitable adoption due to its costs~\citep{aleven_towards_2023}.

Large Language Models (LLMs) have been proposed as one way to scale up access~\citep{aleven_towards_2023}, but significant challenges persist~\citep{miller_llm_2024, macina_opportunities_2023}.
This rise in evidence-based intelligent systems has fueled demand for high-quality educational data.
The few open-source conversational education datasets that exist may not be well-equipped to meet this demand due to their small size~\citep{caines_teacher-student_2020, wang_bridging_2024}, degraded quality from crowdworkers~\citep{yu_burchak_2017, stasaski_cima_2020}, or reliance on LLM tutors or students~\citep{macina_mathdial_2023, miller_llm_2024} which may not be suitable for all downstream tasks~\citep{macina_opportunities_2023, marwala2023algorithm}.
Related, mathematical reasoning is a core challenge in generative AI ~\citep{rane_enhancing_2023}, which has seen an influx of reasoning benchmarks to assess and address this limitation ~\citep{patel_are_2021, li_isarstep_2021, gulati_putnam-axiom_2024}. 
\citet{miller_llm_2024} and \citet{macina_mathtutorbench_2025} benchmarks stand out for their focus on these challenges in the context of LLM tutors.

\begin{figure}[t]
  \centering
  \includegraphics[width=0.8\linewidth]{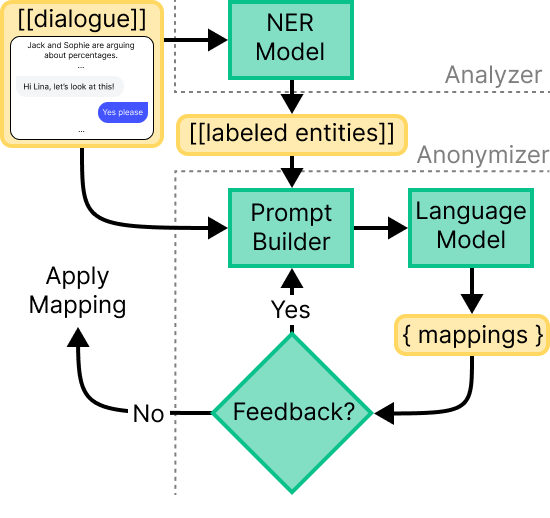}
  \caption{Overview of the PIIvot anonymization framework, which includes a recall-first NER analysis step followed by context-aware surrogate anonymization step.}
  \label{fig:anonymization}
\end{figure}

The sensitive nature of student-generated data presents a significant barrier to sharing real-world educational datasets~\citep{hutt_controlled_2022}.
Frequently, research focuses on personally-identifiable information (PII) as the primary challenge of open-science in sensitive contexts~\citep{olatunji_review_2022}.
Approaches to data anonymization often grow out of healthcare contexts~\citep{olatunji_review_2022} and generally fall into three categories: limiting access, obfuscation, and minimization.
Federated learning limits direct access to records~\citep{antunes_federated_2022, hutt_controlled_2022}, but is not suitable for all types of analysis (e.g., qualitative), and is susceptible to de-anonymization attacks~\citep{carlini_extracting_2021}.
Obfuscating PII typically relies on automated recognition~\citep{buchh_enhancing_2024, bosch_hello_2020, holmes_deidentifying_2023, singhal_-identifying_2024} or manual labeling~\citep{miller_llm_2024}, but identifying PII and overlapping non-PII is challenging even for humans~\citep{singhal_-identifying_2024}.
Finally, both data minimization and k-anonymity aim to reduce the risk of data matching by limiting the exposure to and links between identifiable attributes~\citep{ji_graph_2017, majeed_anonymization_2021, esfandiari_anonymous_2022, sen_diverse_2024, stinar_approach_2024}, but may fall short in contexts where entropy is an important metric of dataset quality~\citep{macina_mathdial_2023}.

Our contribution is two-fold: 1) we developed PIIvot, a novel anonymization framework that reframes PII detection as a simpler potential-PII labeling task and uses an LLM to generate contextually accurate surrogate replacements to preserve data integrity.
Using this method, 2) we open-source a large dataset of question-anchored tutoring dialogues (QATD\textsubscript{2k}) from [a large online math learning platform], demonstrating the effectiveness of PIIvot for anonymizing text-based data at scale.

\section{Method}
\label{sec:method}

\subsection{PIIvot}
Motivated by the high recall of recent PII identification systems and the persistent challenges they face with precision~\citep{buchh_enhancing_2024, bosch_hello_2020, holmes_deidentifying_2023, singhal_-identifying_2024}, we introduce PIIvot, an applied method for text-based anonymization that balances the need to prioritize privacy with data usability.
The framework is grounded in two core principles: (1) a recall-first approach to named entity recognition (NER) for identifying potential-PII (Section~\ref{sec:analysis}), and (2) a Hidden-In-Plain-Sight (HIPS) strategy for generating surrogate replacements that preserve text coherence (Section~\ref{sec:anonymization}). 
This process is illustrated in Figure~\ref{fig:anonymization}.
PIIvot is designed as a generalizable framework that can be adapted to different domains and disclosure risks. 
Here, we detail our specific implementation for transparent data sharing.

\subsubsection{Analysis}
\label{sec:analysis}
The analysis step applies word-level labels to text for named entities that have a risk of containing PII. 
Any suitable NER model can be substituted at this stage, but we caution against openly sharing trained models or open-source details, as they may be used to identify residual PII in the resulting dataset (see \ref{sec:limitations}).
For QATD\textsubscript{2k} we used a DeBERTa model fine-tuned on a prior set of ~40k labeled student-tutor utterances to label dialogue and question text (See Appendix \ref{app:finetune}).~\footnote{We define an utterance as a single chat message where a talk turn can be made up of one or more consecutive messages.}
Specifically, we label names, locations, URLs, date of births, phone numbers, schools, and emails/socials because they are frequent in our data, risk being identifiable, and benefit from granular labels during the anonymization step.
The model applies an IO labeling scheme and first-token aggregation strategy to resolve multi-token predictions into labeled word-level spans. 
Each message is analyzed in a centered context window that includes both the preceding and following messages in the dialogue. 
Finally, we automatically clean labeled spans to remove trailing or preceding punctuation to improve the reliability of downstream surrogate replacement.

\subsubsection{Anonymization}
\label{sec:anonymization}

The anonymization step utilizes labeled spans to generate surrogate replacements under the assumption that the content of non-PII spans can be changed without impacting dataset quality, so long as the same name/location/etc. is consistent throughout the conversation or document.
We argue that this assumption holds for QATD\textsubscript{2k}, where the names and locations of word problems are not relevant to the questions' mathematical concepts. 
This HIPS approach has the added benefit of minimizing the risk of the residual PII problem~\citep{carrell_hiding_2013}.
For labels that can be automatically verified--emails and URLs--we use obfuscation-based anonymization.
For QATD\textsubscript{2k}, we use \textit{GPT-4o-2024-11-20} to generate a mapping from the original set of words to an anonymized set, conditioned on the full chat history to ensure replacements are coherent across each dialogue.
Each label type includes qualities to preserve in the anonymized text that we include in the prompt (i.e., ``When anonymizing [[NAME]], preserve their gender and ethnic background.''). 
Then we apply feedback-based reprompting to enforce measurable qualities of the anonymization (i.e., ensuring the replacement is significantly different from the original).

\subsection{Dataset Collection and Processing}
Existing conversational tutoring datasets ~\citep{macina_mathdial_2023,stasaski_cima_2020,yu_burchak_2017} with annotated talk moves leverage synthetic environments to generate data to scaffold teaching strategies of LLM-based tutors, but limited work has explored these properties in real-world environments.
To fill this gap, we curate a dense set of chat-based tutoring sessions on a UK-based learning platform deployed in over 19,000 schools around the world.~\footnote{\href{https://eedi.com/}{https://eedi.com/}}
Conversations are prompted by the student asking for assistance from an on-demand expert tutor while working on a lesson typically assigned by their teacher.
We include metadata about the question the student was working on and lesson descriptors.~\footnote{Questions were originally presented to students as images. The associated text-based metadata was extracted using the Mathpix API v3, then labeled and validated by tutors.}

\subsubsection{Initial Filtering} 

First, we select conversations that started during a Diagnostic Question (DQ), but before an answer was selected.
DQs are multiple-choice questions with one correct answer and three incorrect distractors representing common misconceptions.
Similarly to ~\citet{chen_predictors_2019}, we filter sessions with at least 20 total and 7 messages from either participant, as these sessions are more likely to have meaningful teaching or learning.
Then, we filter out US-based students by email domain and school.

Next, we take initial steps to safeguard the tutors and students represented in QATD.
We used \textit{omni-moderation-2024-09-26} to filter out conversations with unsafe content.~\footnote{Sexual, sexual/minors, harassment, harassment/threatening, hate, hate/threatening, illicit, illicit/violent, self-harm, self-harm/intent, self-harm/instructions, violence, and violence/graphic.}
We obtained affirmative consent from 25 of 31 tutors represented in the filtered set because of the high density of individual tutors' conversations.
This process resulted in 4,129 dialogues that met our criteria--\textsc{QATD}\textsubscript{Candidate}.

\begin{table*}[t]
\centering
\small
\resizebox{\textwidth}{!}{
\begin{tabular}{lccccccc|cccc}
\toprule
& \multirow{2}{*}{\raisebox{-0.9\height}{\makecell{Total\\Dialogues}}} & \multirow{2}{*}{\raisebox{-0.9\height}{\makecell{Total\\Turns}}} & \multicolumn{2}{c}{Words per Turn} & \multicolumn{2}{c}{N-Gram Entropy} & \multicolumn{1}{c}{\multirow{2}{*}{\raisebox{-0.9\height}{\makecell{Turn\\Uptake}}}} &  & \multicolumn{2}{c}{Human} & \\
\cmidrule(lr){4-5} \cmidrule(lr){6-7} \cmidrule(lr){10-11}
Dataset &  &  & Student & Tutor & Student & Tutor &  & In-Situ & Student & Tutor & Subject\\
\thinmidrule
\thinmidrule
\rowcolor{gray!20}
QATD\textsubscript{2k}\rule{0pt}{2.1ex}& 1971 & \textbf{46249} & 4.15 & 14.79 & 12.74 & 13.39 & 0.69 & \cmark & \cmark & \cmark & Math \\
$\hookrightarrow$\hspace{0.5em}No PIIvot& -- & -- & 4.15 & 14.80 & 12.73 & 13.39 & 0.69 & -- & -- & -- & -- \\
TSCC v2& 260 & 25840 & 9.91 & 18.92 & 13.84 & 14.48 & 0.71 & \cmark & \cmark & \cmark & Lang. \\
Bridge& 459 & 2860 & 2.57 & 14.98 & 10.13 & 11.89 & 0.74 & \cmark & \cmark & \cmark & Math \\
CoMTA& 188 & 2022 & 8.32 & 37.08 & 11.54 & 12.07 & 0.90 & \cmark & \cmark &  & Math \\
CIMA& 391 & 1427 & 6.58 & 10.00 & 8.69 & 10.36 & 0.83 &  & \cmark & \cmark & Lang. \\
Burchak& 173 & 2412 & 3.20 & 3.47 & 10.51 & 10.54 & 0.59 &  & \cmark & \cmark & Lang. \\
MathDial& 2262 & 29453 & 37.86 & 15.88 & 13.82 & 13.79 & 0.84 &  &  & \cmark & Math \\
\bottomrule
\end{tabular}
}
\caption{Comparison of available 1:1 tutoring datasets. Uptake is modeled using~\citet{demszky_measuring_2021}. PIIvot had little to no effect on text-based metrics. }
\label{tab:dataset-overview}
\end{table*}

\subsubsection{Talk Move Downsampling}
Motivated by the growing emphasis on quality over quantity for alignment tasks~\citep{zhou_lima_2023} and data-sharing restrictions, we selectively downsample \textsc{QATD}\textsubscript{Candidate} to create a dataset that prioritizes diverse examples of tutor talk moves.
\textit{Talk moves} are strategies used to support students' mathematical thinking, understanding, and communication~\citep{oconnor_scaling_2015}.
We use the GPT4 talk move classifier from prior work to apply 7 talk move labels~\citep{moreau-pernet_classifying_2024}.
Because this model was fine-tuned on small group tutoring conversations, we first evaluate its generalizability to 1:1 online tutoring.
The first author annotated a weighted stratified sample of 200 tutor messages to conduct a contextual error analysis (see Appendix~\ref{app:talkmove})~\citep{chancellor_contextual_2023}.
Except for a systematic error on the `Getting Students to Relate' label, we see similar performance to the original paper.

To construct our final dataset, \textsc{QATD}\textsubscript{2k}, we first compute TF-IDF scores over talk move labels in \textsc{QATD}\textsubscript{Candidate}, excluding `None' and `GSR'. We form \textsc{QATD}\textsubscript{2k} by greedily selecting dialogues with the max TF-IDF score under two constraints:
(1) at most 8 dialogues per distinct DQ, and (2) a maximum of 1000 unique DQs.
This strategy results in the most diverse examples of tutoring strategies without oversampling from any single DQ.

\subsection{Annotations}

We evaluate the performance of PIIvot on \textsc{QATD}\textsubscript{2k} by manually annotating potential-PII.
A codebook was developed during a prior labeling initiative of ~40k student-tutor messages and achieved a minimum Weighted F1 score of 0.98 between raters across a subset of 350 dialogues (see Appendix ~\ref{app:annotations}).
The first and second authors and two tutors from the original initiative independently applied the codebook to 68,717 messages and 1000 questions.
Discrepancies between the machine and annotator labels were resolved to establish a ground truth.
Each dialogue was also flagged for the presence of unsafe content and the absence of a learning event--29 dialogues were removed, 28 for learning and 1 for safety.

\section{Results and Discussion}
\label{sec:results}

\subsection{PIIvot}
To assess the effectiveness of the PIIvot framework, we triangulate data from curating \textsc{QATD}\textsubscript{2k}. 
We report aggregate label metrics to mitigate the small but non-zero risk of residual PII. 
The high inter-rater reliability observed in the potential-PII labeling task indicates that the task is more straightforward than PII annotation. 
Table \ref{tab:piivot-results} presents macro-averaged metrics comparing our potential-PII NER model against manually annotated labels, evaluated on both dialogue and DQ text.
As expected, our model/task outperforms the more challenging PII detection task on student-generated text in comparable educational domains~\citep{buchh_enhancing_2024}. 
However, we observe degraded performance on LaTeX-formatted DQ metadata (see Section~\ref{sec:limitations}).
Table~\ref{tab:dataset-overview} illustrates that PIIvot anonymization has minimal impact on key text characteristics of \textsc{QATD}\textsubscript{2k}.
These results present a practical case the PIIvot framework in data-sharing pipelines.

\begin{table}[t]
\centering
\small
\begin{tabular}{llccc}
\toprule
\textbf{Label Set} & \textbf{Source} & \textbf{Precision} & \textbf{Recall} & \textbf{F1} \\
\midrule
\multirow{2}{*}{Dialogues} 
  & PIIvot & 0.984 & 0.984 & 0.984 \\
  & Annotators & 0.993 & 0.995 & 0.994 \\
\midrule
\multirow{2}{*}{Questions} 
  & PIIvot & 0.991 & 0.699 & 0.820 \\
  & Annotators & 0.997 & 0.997 & 0.997 \\
\bottomrule
\end{tabular}
\caption{Micro-averaged metrics for potential-PII detection on dialogues and question text compared to ground truth labels.}
\label{tab:piivot-results}
\end{table}

\subsection{QATD\textsubscript{2k}}
\begin{figure}[htbp]
    \centering
    \includegraphics[width=\columnwidth]{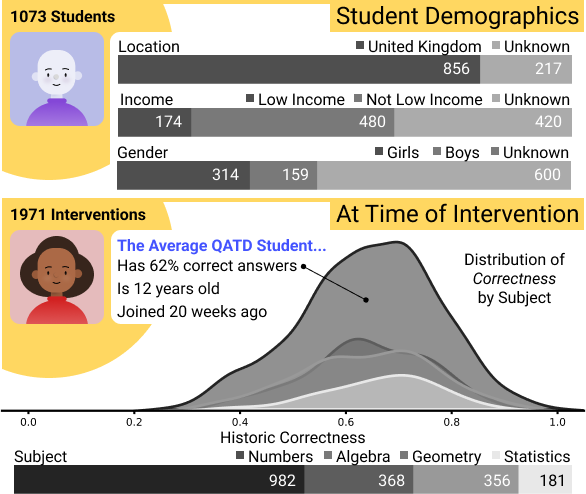}
    \caption{A figure of describing the 1073 students in QATD\textsubscript{2k}. Location, gender, and age are self-reported. The historic correctness plot shows a kernel density estimate (KDE) of student accuracy weighted to prioritize students with 100+ answers.}
    \label{fig:qatd}
\end{figure}

We shift to a brief reflection on QATD\textsubscript{2k}.
Roughly 1\% of sessions were removed due to the absence of a learning event, suggesting talk move downsampling successfully prioritized pedagogically meaningful conversations.
Figure~\ref{fig:qatd} presents an overview of the students represented in the data.
While experiments on QATD\textsubscript{2k} are outside the scope of this work, we provide train/test splits to support comparisons across models and methods in future work.  

Table~\ref{tab:dataset-overview} situates QATD\textsubscript{2k} within the broader landscape of available 1:1 tutoring datasets.
With more real-world data in the available open-sourced tutoring datasets, two trends emerge.
First, LLM tutors/students tend to generate unrealistically long messages.
Second, the high uptake metrics of datasets with synthetic tutors--LLMs or crowdsourcing--indicate potential overfitting to student turns in a way that diverges from authentic responses.
These patterns underscore the importance of real-world tutoring systems to respond effectively in low-information dialogue settings. 
Future work should include more in situ datasets in benchmark and training data preparation.
We open-source QATD\textsubscript{2k} to support this growing demand for real-world tutoring datasets.

\section{Conclusion}
We introduce PIIvot, an anonymization framework that balances the trade-off between precision and recall in PII identification, suitable for contexts where the content of overlapping non-PII entities doesn't impact dataset integrity. 
PIIvot enabled the open sourcing of QATD\textsubscript{2k} to support future research on effective math tutoring.
We present results from curating QATD\textsubscript{2k} as a practical case for using the PIIvot framework in data-sharing pipelines.

\section{Limitations and Ethical Considerations}

\subsection{Limitations}
\label{sec:limitations}

Our work presents two valuable contributions with the PIIvot framework and the QATD dataset, but both carry important limitations that should be considered in future research and downstream applications.

We acknowledge that the PIIvot framework relies on the assumption that the content of labeled entities is insignificant, which is not true across many domains.
Future work could explore improved prompting strategies and/or feedback during anonymization to better preserve the significance of replaced content and mitigate this limitation in new contexts.
Recent work demonstrates the potential of incorporating LLM-generated feedback to improve LLM summarization tasks~\citep{song_learning_2025}, suggesting a promising direction for PIIvot anonymization feedback.
Related, the framework relies on an effective NER that meets the privacy needs of one's data.
Additionally, PIIvot uses HIPS to obfuscate PII.
We strongly recommend that neither the underlying labels nor the NER models be released alongside datasets, as they may expose residual PII.
In our case, there still remains a non-zero risk of residual PII in QATD, despite extensive measures to ensure the safety and privacy of tutors and students.
This risk illustrates an inherent limitation of any automated anonymization pipeline and underscores the need to consider a variety of privacy factors outside of identifiably. 

Related to the QATD dataset, we highlight four key limitations that reflect trade-offs made to support open-source release and downstream usability.
First, anonymization of DQs in QATD relies heavily on human annotation due to poor generalizability of our NER model to this text format.
We accept this limitation because question text has no privacy risks and is easier to label due to its limited volume and predictable format.
Second, we acknowledge that the `Getting Students to Relate' talk move label may not fully generalize to our 1:1 tutoring context.
We include talk move labels in QATD for method transparency, but downstream use of this signal should consider this limitation (Discussed further in Appendix~\ref{app:talkmove}).
Third, this dataset reflects real interactions on Eedi, where tutors occasionally manage multiple students during peak hours and prioritize resolving misconceptions to help students feel confident getting back to their lesson.
This context and the reported behavior and demographic factors in Figure~\ref{fig:qatd} should be considered when interpreting tutor and student behavior in QATD.
Finally, we acknowledge that our decision to prioritize student privacy by removing student links across tutoring sessions may impact downstream applications of QATD.
This decision was made due to the inability to get additional student consent outside of the platform's terms and conditions for the risks conversation linkage could introduce.
We underscore that PII anonymization is only one aspect of responsible data sharing and broader privacy concerns.

\subsection{Ethical Considerations}
This work highlights the range of privacy considerations necessary when open-sourcing data from real educational platforms.
While this work is outside the purview of what is traditionally defined as human subjects research, we recognize our responsibility to reflect on its ethical implications--both for dissemination and shaping best practices for future research.

First, Eedi’s legal terms of service and privacy policy permit the sharing of personal data with third parties for the purpose of conducting research, but we recognize that legal permission alone is not sufficient.
Prior research emphasizes the ethical responsibility of researchers and platform organizers to steward the trust of their users/stakeholders~\citep{commission_protecting_nodate, zent_beyond_2025}.
Considering these values, we obtained affirmative consent from high-volume contributors, applied data minimization principles to student data, and outline the following recommendations for appropriate secondary use.
In accordance with Eedi's privacy policy, QATD is released for non-commercial research (under cc-by-nc-sa-4.0) aimed at improving student learning outcomes, including, but not limited to, dialogue modeling, model calibration, and tutoring interaction analysis.
Attempts to re-identify individuals from QATD are out of scope and violate the intended use of this dataset.
We encourage future research to use this dataset to advance understanding of how conversational strategies support learning while upholding these ethical standards.

We further caution researchers to validate third-party APIs used in PIIvot anonymized to ensure prompt inputs are not stored or logged, as they contain unanonymized text.
In our case, OpenAI reports not using our prompt data for model training or persistent storage.
We encourage future work to consider self-hosting LLMs for highly sensitive contexts.

Finally, we acknowledge the positionally of the authors and annotators of this work as paid employees of Eedi.
This relationship carries both privileged access and heightened ethical responsibility. 
As stewards of users’ trust, our proximity to the platform and its data influenced our anonymization decisions. 
We prioritized safety and privacy, opting for conservative redaction and aggregation strategies and human validation to minimize the risk of re-identification. 
This commitment reflects our obligation to protect the individuals whose interactions make this research possible.

\subsubsection{AI Assistant Disclosure}
We used AI assistants, including Copilot and GPT, to support code development and documentation. 
We used these tools to draft boilerplate code and text for some comments and documentation. 
All generated content was validated and iterated on to align with our standards.

\section*{Acknowledgments}
We thank Eedi for supporting this work and the committed tutors whose dedication made this research possible. 
We are especially grateful to those who contributed their time and expertise to the annotation process.

\bibliography{bibliography, anthology}

\appendix

\section{Potential-PII NER Model}
In this section, we describe our process for NLP model fine-tuning on the potential-PII classification task. 
We developed our own classification model for two reasons: 1) initial exploration of existing PII identification models revealed poor performance on UK names, and 2) we wanted more control of label granularity to support surrogate replacement.
First, we outline the annotation process to support supervised fine-tuning, and then we discuss our experimental setup and hyperparameters.

\begin{table}[h]
\centering
\begin{tabular}{l c}
\toprule
\multicolumn{2}{c}{\textbf{DeBERTa-PIIvot-NER-IO}} \\
\midrule
Precision & 0.93 \\
Recall & \textbf{0.98} \\
F1 & 0.94 \\
Balanced Accuracy & 0.98 \\
\bottomrule
\end{tabular}
\caption{Performance of the final DeBERTa-PIIvot-NER-IO model on a held-out test set. Macro scores are computed over positive labels; balanced accuracy includes the `O' (non-PII) class.}
\label{tab:pii_model_performance}
\end{table}

\subsection{Potential-PII Annotation}
\label{app:annotations}

\begin{table*}[t]
\centering
\small
\begin{tabular}{llccc}
\toprule
& & \multicolumn{3}{c}{\textbf{Approximate Support}} \\
\cmidrule(lr){3-5}
\textbf{Label} & \textbf{Description} & \textbf{Train} & \textbf{Validation} & \textbf{Test} \\
\midrule
I-date\_of\_birth & Birth date detail & 90 (94\%) & 20 (95\%) & $<$10 (0\%) \\
I-email\_social & Email address, social media handle, or profile & 80 (92\%) & 20 (95\%) & $<$10 (0\%) \\
I-location\_address & Geographical detail indicative of a person's location & 100 (65\%) & 30 (69\%) & 10 (0\%) \\
I-name & A person's full, partial, or nickname & 2300 (0\%) & 600 (0\%) & 700 (0\%) \\
I-phone\_number & Phone number & 80 (96\%) & 20 (95\%) & $<$10 (0\%) \\
I-school\_name & School name & 70 (95\%) & 20 (94\%) & $<$10 (0\%) \\
I-url & URL & 100 (58\%) & 30 (68\%) & 20 (0\%) \\
\bottomrule
\end{tabular}
\caption{IO label schema and approximate support (with \% synthetic) in each dataset split.}
\label{tab:pii_distribution}
\end{table*}

\begin{table*}[t]
\centering
\small
\resizebox{\textwidth}{!}{
\begin{tabular}{lcccccc}
\toprule
\textbf{Label} &
\makecell[c]{\textbf{Support}\\\textbf{QATD\textsubscript{Candidate}}} &
\makecell[c]{\textbf{Support}\\\textbf{Validation Set}} &
\makecell[c]{\textbf{Support}\\\textbf{Original}} &
\textbf{F1} &
\makecell[c]{\textbf{F1 Excl.}\\\textbf{\textless GSR\textgreater}} &
\makecell[c]{\textbf{F1}\\\textbf{(Original)}} \\
\midrule
\textless None\textgreater & 0.42 & 0.28 & 0.73 & 0.8991 & 0.8991 & \textbf{0.96} \\
\textless Keep Together\textgreater & 0.36 & 0.20 & 0.09 & 0.8333 & \textbf{0.8750} & 0.81 \\
\textless Revoicing\textgreater & 0.12 & 0.18 & 0.03 & 0.8986 & \textbf{0.8986} & 0.76 \\
\textless Press for Accuracy\textgreater & 0.06 & 0.16 & 0.13 & 0.7733 & 0.8286 & \textbf{0.88} \\
\textless Getting Students to Relate\textgreater & 0.02 & 0.08 & 0.004 & 0.0000 & -- & \textbf{0.75} \\
\textless Press for Reasoning\textgreater & 0.002 & 0.06 & 0.006 & 0.7857 & \textbf{0.9565} & 0.94 \\
\textless Restating\textgreater & 0.0003 & 0.04 & 0.008 & 0.8000 & 0.8000 & \textbf{0.95} \\
\bottomrule
\end{tabular}
}
\caption{Distribution and F1 scores for talk move labels comparing our dataset with the original metrics in ~\citet{moreau-pernet_classifying_2024} F1 scores reported with and without the \textless GSR\textgreater\ label.}
\label{tab:talk_move_f1_comparison}
\end{table*}

Annotation for an independent batch of 66,059 tutor/student messages took place from May to August 2024 by paid employees of Eedi.
The first and second authors (US/UK/Male) and 4 expert tutors (UK/Female) participated in this annotation process; tutors annotated messages while not actively helping students.
We used the open-source annotation tool Doccano
to apply labeled spans to tutor messages~\citep{doccano}.
Before manual annotation, labels are prepopulated using a regex applied matching strategy using the known first and last names of the tutor and student, as well as common word problem names.

First, we developed and validated a codebook to support potential-PII labeling.
Annotators independently labeled a subset of 350 dialogues, achieving a minimum Weighted F1 score of 0.98 between raters.
This high level of agreement indicated that the codebook was well calibrated, and no significant changes were needed.
The codebook, including annotation instructions, is available in the PIIvot repository.
Annotators individually applied the codebook to the remaining messages to support model fine-tuning.

\subsection{Model Fine-Tuning}

\label{app:finetune}

We conduct our model fine-tuning experimentation on a single NVIDIA Tesla V100 GPU using deberta-v3-base (184M parameters) and bert-base-uncased (110M parameters) ~\citep{he2021debertav3, DBLP:journals/corr/abs-1810-04805}.
To support model testing, we use stratified sampling on the minority label for a given message to generate train (64\%), test (20\%), and validation (16\%) splits (see Table~\ref{tab:pii_distribution} for approximate label splits).
We initialize a sequential set of hyperparameter grid searches over a select subset of approaches.
This greedy approach allowed us to explore a wide variety of modeling approaches without ballooning compute time.

Optimal configurations are shown in bold.
For each search, we also include two learning rates ($1\text{e}{-5}$ vs. \textbf{$\boldsymbol{2\text{e}{-5}}$}) and two labeling schemas (\textbf{IO} vs. IOB2) over 4 epochs with early stopping on performance degradation.
In order we test BERT vs. \textbf{DeBERTa}, Adam vs. \textbf{AdamW}, \textbf{raw + synthetic} vs. raw data, and \textbf{windowing} vs. non-windowing.
For the synthetic data, 20 PII-rich synthetic student-tutor conversations were manually created by the authors to augment the training and validation data with examples of imbalanced classes.
For the windowing condition, model inputs include the prior and preceding message.
In total, fine-tuning and final model training required 36 GPU hours.
 
The final model was trained using the AdamW optimizer with $\beta_1{=}0.9$, $\beta_2{=}0.999$, $\epsilon{=}1\text{e}{-8}$, and a weight decay of 0.01. 
We used a learning rate of $2\text{e}{-5}$, a batch size of 4, and trained for 3 epochs with a random seed of 42.
We report model performance on the hold-out test split in Table~\ref{tab:pii_model_performance}.

\section{Talk Move Classification}

\label{app:talkmove}

To facilitate downsampling, talk moves were applied using the GPT-based classification model in \citet{moreau-pernet_classifying_2024}. 
Labels include: `Pressing for accuracy', `Keeping everyone together',`Revoicing', `Restating', `Pressing for reasoning', `Getting students to relate to another’s ideas', and `None'.
The model was fine-tuned on conversation transcripts from small-group math tutoring sessions.
Both sets of authors decided use of this artifact was acceptable so long as performance was validated in this new context.

\subsection{Contextual Error Analysis}
To evaluate the generalizability of the model for 1:1 chat-based math tutoring sessions, the first author manually annotated a validation set of 200 tutor utterances sampled through weighted stratified sampling using the original codebook of \citet{moreau-pernet_classifying_2024}. 
The sample distribution was flattened by 0.8 of the original label distribution represented in QATD\textsubscript{Candidate} in order to validate more examples of minority class labels (see Table~\ref{tab:talk_move_f1_comparison}).

The first author conducted a contextual error analysis on all mismatched labels~\cite{chancellor_contextual_2023}.
This method introduces qualitative coding and thematic analysis into traditional ML error analysis to understand contextual details missed in annotation tasks.
We adopt contextual error analysis for this task because it is well-equipped to reveal aspects of the model that don't generalize to 1:1 tutoring contexts. 

We begin by qualitatively coding tutor messages and memoing contextual errors. 
Two themes emerged from these artifacts that we use to describe the errors introduced by applying the model to this new context.
First, a small source of errors related to the \textbf{multi-message chat turns} present in QATD. 
When tutors span their intent across multiple messages, the temporal fragmentation leads to label mismatches or partial crediting of complex moves. 
A major class of errors stemmed from the \textbf{fictional argument questions} used in DQs.
These items frame math problems as debates between two fictional students, and tutors frequently probe the student to reason about the validity of each claim.
While these prompts closely resemble <Getting Students to Relate> (<GSR>) in structure, the original codebook doesn't take a stance on whether this label applies to a fictional setting.
We chose not to apply the <GSR> to these instances, but acknowledge this is a gray area for a clearly out of context example.
We note that in all cases, <GSR> was associated with another positive talk move label.
We use these observations to motivate reporting a second set of metrics excluding the controversial label.
We find that our results are comparable to those reported in~\citet{moreau-pernet_classifying_2024} and use this as grounds for applying the classifier to downsample QATD\textsubscript{Candidate} to QATD\textsubscript{2k}.

\end{document}